\documentclass[conference]{IEEEtran}
\IEEEoverridecommandlockouts
\usepackage{cite}
\usepackage{amsmath,amssymb,amsfonts}
\usepackage{algorithmic}
\usepackage{graphicx}
\usepackage{textcomp}
\usepackage{xcolor}
\definecolor{darkgreen}{RGB}{15, 127, 4}
\usepackage[super]{nth}

\def\BibTeX{{\rm B\kern-.05em{\sc i\kern-.025em b}\kern-.08em
    T\kern-.1667em\lower.7ex\hbox{E}\kern-.125emX}}
\begin{document}

\title{DroTrack: High-speed Drone-based Object Tracking Under Uncertainty
}

\author{\IEEEauthorblockN{
% 1\textsuperscript{st} 
Ali Hamdi}
\IEEEauthorblockA{RMIT University \\
Melbourne, Australia. \\
ali.ali@rmit.edu.au}
\and
\IEEEauthorblockN{
% 2\textsuperscript{nd} 
Flora Salim}
\IEEEauthorblockA{RMIT University \\
Melbourne, Australia. \\
flora.salim@rmit.edu.au}
\and
\IEEEauthorblockN{
% 3\textsuperscript{rd} 
Du Yong Kim}
\IEEEauthorblockA{RMIT University \\
Melbourne, Australia. \\
duyong.kim@rmit.edu.au}
}

\maketitle

\begin{abstract}
We present DroTrack, a high-speed visual single-object tracking framework for drone-captured video sequences. Most of the existing object tracking methods are designed to tackle well-known challenges, such as occlusion and cluttered backgrounds. The complex motion of drones, i.e., multiple degrees of freedom in three-dimensional space, causes high uncertainty. The uncertainty problem leads to inaccurate location predictions and fuzziness in scale estimations. DroTrack solves such issues by discovering the dependency between object representation and motion geometry. We implement an effective object segmentation based on Fuzzy C Means (FCM). We incorporate the spatial information into the membership function to cluster the most discriminative segments. We then enhance the object segmentation by using a pre-trained Convolution Neural Network (CNN) model. DroTrack also leverages the geometrical angular motion to estimate a reliable object scale. We discuss the experimental results and performance evaluation using two datasets of 51,462 drone-captured frames. 
The combination of the FCM segmentation and the angular scaling increased DroTrack precision by up to $9\%$ and decreased the centre location error by $162$ pixels on average.
DroTrack outperforms all the high-speed trackers and achieves comparable results in comparison to deep learning trackers. DroTrack offers high frame rates up to 1000 frame per second (\textit{fps}) with the best location precision, more than a set of state-of-the-art real-time trackers. 
% DroTrack ranks second and third (out of 10 trackers) in most of benchmarking experiments.

\end{abstract}


\section{Introduction}

Drone-related research has been widely pursued over the past several years. Drones are aerial platforms with advanced equipment, e.g., high-resolution cameras. They offer low-cost, safe operations to monitor locations inaccessible to humans. 
Classical imagery devices, such as satellite and street-level cameras, suffer from various limitations such as low resolution and low detail, respectively. Drones fly at low altitudes and offer a wide field of view to capture high-resolution images and greater detail. 
Visual drone applications include topographic mapping, surveillance, and search and rescue.
However, drone-based video quality is affected by different uncertainties. Drones' motion is dependant on multiple situation inputs such as the weather conditions and structures of tracking locations.

Visual object tracking is a key component of the aforementioned drone applications. It has many challenges, such as noise, occlusion, cluttered backgrounds, and object varying features \cite{apeltauer2015automatic}. 
The intrinsic variability in object's colour or shape causes poor tracking predictions. This challenging issue is caused by the uncertainty of tracking environments' aspects. For example, occlusion can occur due to the shadows of trees and buildings. 
In addition, drones move with multiple degrees of freedom in three-dimensional space.
This leads to unexpected changes in the drone and object locations pose multiple uncertainty and fuzziness issues in the object rotation and scale \cite{moranduzzo2014automatic}.
The point tracking algorithms depend on one or more features, i.e., key-points or corners. An incremental shift may occur in the tracking location due to the spatial distances between the correct and predicted tracking points. Thus, the performance of the existing tracking algorithms is degraded in different drone-based tracking situations. Moreover, drone-based object tracking requires real-time tracking. However, most of the recent trackers utilise deep learning to achieve high-accuracy tracking regardless of the tracking speed \cite{bertinetto2016fully}. In this paper, we propose a robust object tracking algorithm that discovers the relationship between the object's visual and geometrical representations to overcome such challenges in real-time.

\begin{figure*}[!h]
\begin{center}
   \includegraphics[width=0.99\linewidth]{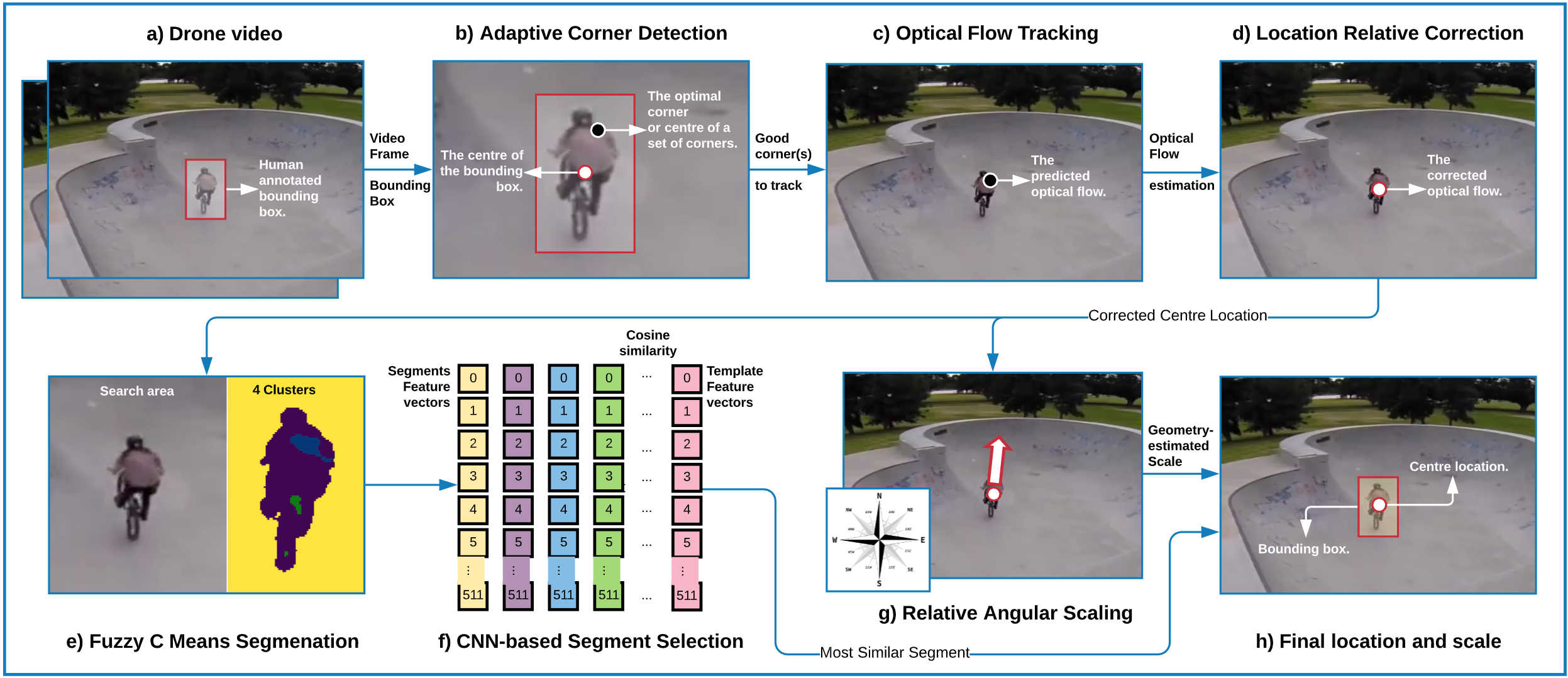}
\end{center}
   \caption{DroTrack includes multiple components for single object tracking, as follows: 
   a) reading the drone-captured video and the bounding box of the object at the first frame are given; 
   b) detecting the optimal corner(s) to be tracked; 
   c) estimating the optical flow; 
   d) correcting the optical flow performance; 
   e) segmenting using fuzzy c means clustering; 
   f) using a VGG16 pre-trained model for extracting convolutional features and comparing the similarity between the feature vectors of the original reference template and the different segments to select the best one; 
   g) calculating the relative angular scaling; 
   and h) incorporating the outcome of (f) and (g) to produce the final tracking location centre and scale.}
   \label{DroTrackFCM}
\end{figure*}

The main contribution of this paper is to solve the uncertainty problem in drone-based single object tracking. DroTrack tackles the impacts of object representation and geometrical drone motion uncertainties on the tracking location and scale. Fig. \ref{DroTrackFCM} illustrates the different components of our methodology. The proposed DroTrack makes the following contributions:
\begin{itemize}
\setlength\itemsep{0em}
\item Adaptive feature extraction and optical flow methods that produce real-time single object tracking.
\item A spatial segmentation method that incorporates a Fuzzy C Means clustering algorithm with a pre-trained CNN transfer learning model.
\item A heuristic geometrical method to estimate accurate object scales.
\item Comprehensive evaluation and benchmark with the baseline and state-of-the-art trackers using two drone-captured datasets with 51,462 frames.
\end{itemize}

The structure of this paper is as follows. We discuss the related works and the problem scenario in Sections \ref{rw} and \ref{ps}, respectively. The research methodology underlying the DroTrack components is explained in Section \ref{sec:DroTrack}. Section \ref{exp_results} presents the experimental results and evaluations, and Section \ref{conclusion} concludes the work.

%-------------------------------

%------------------------------------------
\section{Related Work}\label{rw}
The main goal of visual object tracking is to discriminate an object's area from the background in a sequence of frames. Tracking concerns the estimation of the location and scale of the object in each frame. In the literature, there are two main streams for object tracking methods, based on appearance and motion. Recently, object tracking has been studied in multiple studies, such as \cite{li2017learning,zhang2017multi,mueller2017context}. Much research is available on fixed camera scenarios, whereas only limited research can be found on moving camera situations \cite{li2017visual}.

There are a few studies dedicated to drone-based object tracking. The work by \cite{zhang2017visual} utilised fast feature pyramids and a median flow tracker for pedestrian detection and tracking. Multiple hypothesis tracking (MHT) was used for multi-object tracking \cite{jianfang2017novel}. The MHT-based framework is based on a multidimensional assignment formulation using a time-slide window approach. The work by \cite{gaxiola2016target} developed a composite correlation filter that is adaptively tuned to recognise the object of interest. The authors in \cite{zhang2016gm} implemented a multi-object Bayesian filter based on probability hypothesis density approximation. The implemented multi-object filter revises the weights of close tracking targets and reduces the disturbance of clutter. An online drone-based object tracking controller is developed in \cite{kendall2014board}. The proposed system tracks an object of predefined colour without external localisation sensors or GPS. This system corrected the predicted motion using a Kalman filter.

On the other hand, deep learning networks have been widely utilised in various computer vision applications including moving object tracking \cite{bertinetto2016fully,li2019target,choi2019deep,danelljan2015convolutional,wang2015transferring}. The major problem is the limited knowledge that available to train the deep networks online to track objects that was only seen previously in one example. One possible solution is to train deep CNN for object tracking. However, the lack of annotated data hinders the training of the deep CNN. Training an offline CNN using a large set of videos with tracking ground-truths can solve this issue by transferring the learned features hierarchies to online tracking \cite{WangLGY2015Transferring,NamH2015Learning}. Moreover, Graph Convolutional Networks are also proposed for object tracking \cite{gao2019graph}. The main issue of these methods is to perform Stochastic Gradient Descent online to adapt the weights of the network. This requires high computational speed which makes the tracking unreliable, especially in the drone scenario.

The existing trackers often fail for high-speed objects and unmodeled drone motion. For example, the directions of such high-speed objects can easily be changed by 360 degrees. 
The use of traditional trackers is also affected by their low-speed tracking, which is not preferred in drone scenarios. Most existing research in drone-based tracking depends on either appearance \cite{kendall2014board} or feature point detection and tracking \cite{nussberger2014aerial}. We design our solution to consider new uncertainties in the drone-based object tracking. To produce an accurate drone-based object tracker, we propose to enhance the appearance-based tracking and use the motion characteristics of the object to calculate its relative scale. 

%-------------------------------------------------------------------------

\begin{figure}[t]
\begin{center}
  \includegraphics[width=.85\linewidth]{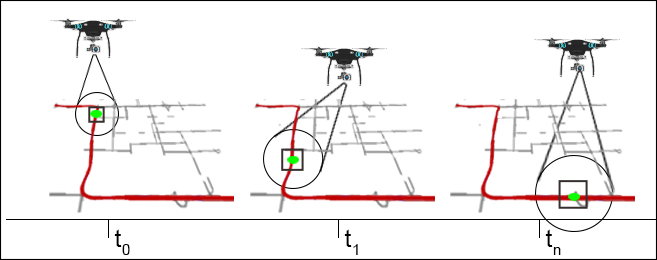}
\end{center}
   \caption{Object locations and scales at two different time-stamps.}
\label{problem-scenario}
\end{figure}
\section{Problem Formulation and Scenario}\label{ps}
\begin{figure}[ht]
\begin{center}
   \includegraphics[width=0.7\linewidth]{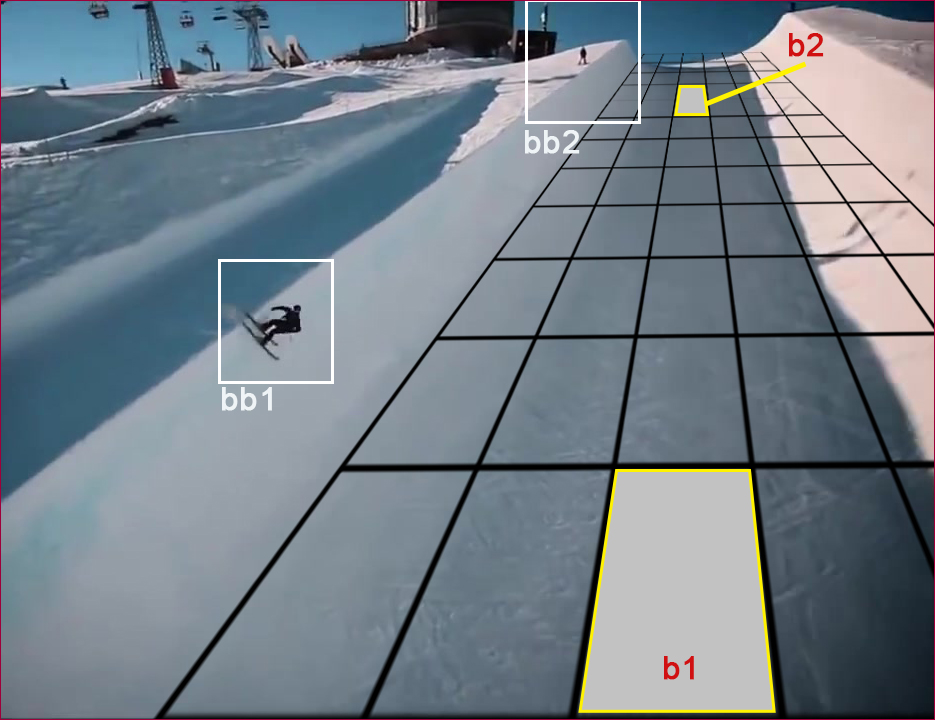}
\end{center}
   \caption{An example of a planar scene.}
\label{planar}
\end{figure}

We formulate a drone-based tracking problem as follows. A drone \emph{d} is tracking a moving \emph{o} in real time using a camera as illustrated in Fig. \ref{problem-scenario}. Unlike conventional object tracking using fixed cameras, a camera mounted on \emph{d} is moving according to the motion of \emph{d}.
When \emph{d} or \emph{o} moves the distance between them is altered. This leads to changes in the location and scale of \emph{o} in the video frame. 
We propose to solve these uncertainties issues using visual and geometrical reasoning.
Fig. \ref{problem-scenario} shows three different tracking positions of a drone in different time-stamps. The drone monitors a moving object indicated in light green. As illustrated, the scale of the moving object is inversely related to the size of the drone's field of view. When the drone flies high and has a wide field of view, the object becomes smaller. Conversely, the object scale is enlarged if the drone becomes close.
Fig. \ref{planar} shows the planar projection issue on two objects in the same scene. The bounding boxes \emph{bb1} and \emph{bb2} contain two men located at two different depths, i.e., distances from the drone camera. Their different depths affect their projections, which have different pixel representations in the image frame. Moreover, the boxes \emph{b1} and \emph{b2} represent two equal blocks on the grid, although their visual representations are varied. As will be discussed in Section \ref{exp_results}, the evaluation of DroTrack is done using two datasets that involve video segments captured in different environments with different scene structures and object depths.

We propose to track the object's location and infer its scale based on the variation of its visual representation and motion features. Using a Fuzzy based segmentation methodology helps to locate the object accurately. The accurate computing of geometrical relationships between the moving object and drone allows the tracking framework to ensure correct long-term tracking.

%--------------------------------------------------------------
%
% \begin{figure}[ht] % figure*
% \begin{center}
%   \includegraphics[angle=0, origin=c, width=0.9\linewidth]{images/Drone-tracking-2}
% \end{center}
%   \caption{DroTrack implemented components.}
% \label{DroTrack}
% \end{figure}
%%--------------------------------------------------------------
\section{DroTrack: Drone-based Object Tracking}\label{sec:DroTrack}
DroTrack, the proposed drone-based object tracking method, has five main components, as showed in Fig. \ref{DroTrackFCM}, as follows:
1) adaptive corner detection,
2) fast single-point optical flow tracking,
3) optical flow relative correction,
4) Fuzzy C Means based segmentation.
4) angular relative scaling.
The feature-based tracking and intelligent scale estimation offer very high-speed tracking without compromising its accuracy. In the following sub-sections, we explain each component.
\subsection{Adaptive Corner Detection Algorithm}\label{acda}
\begin{figure}[t]
\begin{center}
   \includegraphics[width=0.85\linewidth]{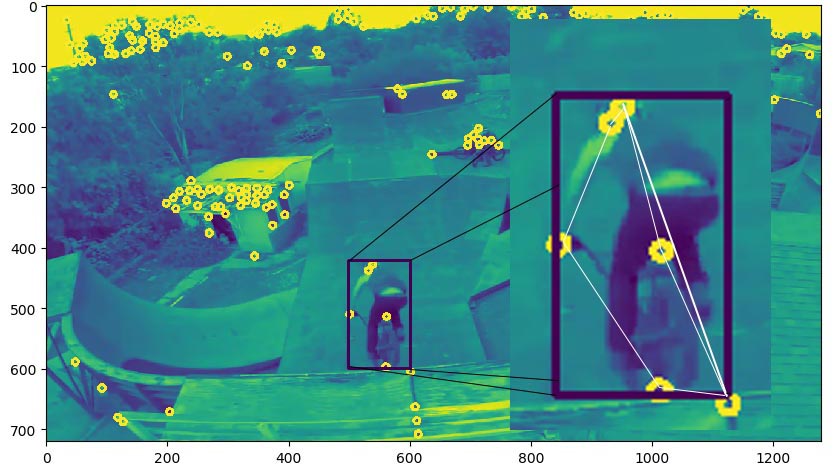}
\end{center}
   \caption{An example of the the convex hull of multiple corners.}
\label{multi-points}
\end{figure}

A \emph{corner} is a point-of-interest represented as an image pixel where any detected edge changes its direction significantly in two dimensions. Corners offer a better choice for object tracking. They enable tracking changes in two dimensions that cannot be detected with other features, such as edges, where the changes are only in one dimension \cite{Juranek-2018}.
DroTrack detects the strong corners on the first frame (FF) using the Shi Tomasi method \cite{shi1994good}. The best corners are selected based on their Shi Tomasi parameters (STPs), such as quality level, minimum distance, a derivative covariation matrix block size and a maximum number of corners. The Shi Tomasi method is an extension of the Harris Corner Detector (HCD). The HCD enables invariance to rotation, scale, illumination variation and noise. It utilises a local auto-correlation function that employs small shifts to measure local changes in different directions.
Here, DroTrack focuses on the closest corner(s) to the centre of the reference template (RT), i.e., the image segment that falls in the human-annotated bounding box at the first frame. There maybe one or more closest corners detected inside the RT. Fig. \ref{multi-points} shows an example of detected multiple corners (coloured in yellow) and the connecting convex hull (coloured on white). For simplicity, in the rest of this paper, closest corner (s) \textbf{CC} is used to represent one or multiple selected corners.
In order to achieve this, we propose an adaptive algorithm that works recursively to find the CC. The algorithm begins the corners detection with STPs of high quality, small corner number, distance threshold and block size. The algorithm decides either to continue tracking these corner or to tune the STPs and redo the previous process until the best CC is found based on two evaluation scores: similarity and distance.

We define a test template to be the area positioned around the detected corner, with the same width and height of the RT. In case of multiple corners, the centre of their convex hull is used as the template centre, see Fig. \ref{multi-points}. Then we extract the histograms, i.e., graphical representations of the tonal distribution of the image pixel intensity values, of both the RT and the test template. The similarity ratio (\textit{simi.}) is computed as a correlation score between the histograms of the RT and the test template, both in colours. We use the correlation measured according to Eq. \ref{eq:hist_corr} to compute the correlation ratio between the two histograms. The score is between 0 and 1 for the lowest and highest similarities, respectively. If the \textit{simi.} is below a certain threshold ($\alpha$), the algorithm will adapt the feature extraction parameters to achieve a more fine-grained search. 
\begin{equation}\label{eq:hist_corr}
    simi.(H_1, H_2) = \frac{\sum_I (H_1(I) - \bar{H}_1)(H_2(I) - \bar{H}_2)}{\sqrt{\sum_I (H_1(I) - \bar{H}_1)^2 \sum_I (H_2(I) - \bar{H}_2)^2}}
\end{equation}
Where \emph{$H_1$} and \emph{$H_2$} are the histograms of the RT and test template, \emph{I} iterates for each histogram bin, and the $\bar{H}$ refers to the histogram mean.
The distance between the RT centre point and the CC coordinates is computed. DroTrack relates the distance threshold ($\beta$) to the object scale. It also requires the CC to be less distant than the distance-scaled threshold ($\beta$). 
Using the distance formulas in Eq. \ref{eq:hist_corr} and \ref{eq:dist} based on the histogram correlation and Pythagorean theorem, respectively, 
the function \emph{f} in Eq. \ref{eq:f} compares the \emph{simi.} and \emph{dist.} for the given corner.  Eq. \ref{eq:f} returns 1 if the corner is selected or 0 to rerun the algorithm. Here, $\alpha$ and $\beta$ represent the similarity and distance thresholds and \emph{C} is the given corner. The threshold of the \emph{simi.} ($\alpha$) is defined as 0.5 of the histogram similarity and the \emph{dist.} ($\beta$) is defined as 0.5 of the sum of boundary box’s height and width.

\begin{equation}\label{eq:dist}
    dist. ((x_1,y_1),(x_2,y_2)) = \sqrt{(x_2-x_1)^2+(y_2-y_1)^2}
\end{equation}
where, $x_1,y_1$ and $x_2,y_2$ represent the coordinates of the RT centre and the given corner, respectively.
\begin{equation}\label{eq:f}
    f(C)= 
\begin{cases}
    1,& \text{if } simi. \geq \alpha~~ \&~~ dist. \leq \beta\\
    0,              & \text{otherwise}
\end{cases}
\end{equation}

Specifically, the adaptive corner detector uses a set of minimum and maximum thresholds that are used to match the \textit{simi.} and \textit{dist.} So, the algorithm keeps iterating on the given threshold range until it meets the \textit{simi.} and \textit{dist.}.
The Open CV library implements the Shi Tomasi method with a region of interest parameter, whereas the proposed adaptive method is well designed to enhance the performance of the algorithm by starting at higher quality and with fewer corners. 
In addition, the proposed algorithm overcomes the limitation of the dependency of corner-based tracking on the environmental factors.

\subsection{Optical Flow Tracking}
The selected CC is used to develop the optical flow tracking method. DroTrack employs the implementation of a sparse iterative Lucas-Kanade optical flow in pyramids \cite{bouguet2003pyramidal}. The optical flow method takes two consecutive frames, and a set of corner coordinates belong to the previous frame (PF). In our case, DroTrack only passes the CC coordinates to the optical flow method and obtains the new coordinates. 
\subsection{Optical Flow Relative Correction}\label{ofrc}
We consider the distance between the CC (the closest corner or the convex hull centre) and the FF centre point as a correction margin; see $\Delta_{x}$ and $\Delta_{y}$ in Eq. \ref{cx_cy} and Fig. \ref{DroTrackFCM} at (b). The new coordinates are calculated based on the relative scale of the new RT; see $RT_{scale_{F}}$ in Eq. \ref{scale_f}. 
\begin{equation}\label{cx_cy}
    \Delta_{x} = FF_x - F_x, \Delta_{y} = FF_y - F_y,
\end{equation}
\begin{equation}\label{scale_f}
    RT_{scale_{FF}} = \frac{h_{RT_{FF}}}{h_{FF}}, RT_{scale_{F}} = \frac{h_{RT_{F}}}{h_{F}}
\end{equation}
where $h$ refers to the height.

Thus, the x and y coordinates of corrected point (CP) are computed as in Eq. \ref{cp_x} and Eq. \ref{cp_y}, respectively.
\begin{equation}\label{cp_x}
    CP_{x} = F_x + \Delta_{x} * \frac{RT_{scale_{F}}}{RT_{scale_{FF}}} 
\end{equation}
\begin{equation}\label{cp_y}
    CP_{y} = F_y + \Delta{y} * \frac{RT_{scale_{F}}}{RT_{scale_{FF}}}
\end{equation}
%
% \begin{figure}[t]
% \begin{center}
%   \includegraphics[width=0.6\linewidth]{images/correction}
% \end{center}
%   \caption{Optical flow relative correction based on the distance between the \emph{RT} centre and the closest corner.}
% \label{correction}
% \end{figure}
% %
The relative correction is useful due to the fact that whenever the object moves away from the camera its size changes and has different location centre. The output corrected location centre is used later by the Fuzzy-based segmentation and geometrical algorithms.

\subsection{Object Segmentation with Fuzzy C-means}
In some cases, DroTrack loses tracking of the object due to the uncertainty or fuzziness of the bounding area around the object. Therefore, we propose to apply object segmentation based on a Fuzzy C Means (FCM) methodology. The conventional FCM is sensitive to the image noise. Basic FCM algorithm expects the data to have separate clusters in order to produce accurate membership values. However, this dependency on the cluster similarity is not suitable for image data. This is because the neighbour clusters in an image are highly correlated. Multiple research works propose to overcome this problem, such as \cite{chuang2006fuzzy}. They harness the spatial information to overcome the sensitivity issue of FCM. They simply compute the likelihood that a neighbourhood pixel belongs to a certain cluster. Then, the spatial likeliness score is injected into the membership function. We employ the methodology presented in \cite{tripathy2014image}. They propose to incorporate the hesitation degree and spatial likeliness in the membership function calculated as (msh) in Eq. \ref{msh}.
\begin{equation}\label{msh}
msh_{ij}=\frac{u_{ij}^{p}h_{ij}^{q}}{\sum\limits_{k=1}^{c}u_{kj}^{ p}h_{kj}^{q}}
\end{equation}
where $msh_{ij}$ represents the membership values of the given neighbourhood pixel, $i$ and $j$ represent the pixel coordinates, $u$ is the membership function computed with hesitation score, $h$ is the spatial function, and $p$ and $q$ control the weights of the initial membership and spatial functions, respectively. 
We then apply morphological transformation methods, including erosion and dilation, to clean the noise after segmentation. 
We decide to segment the search area into n clusters, e.g., 5. 

\begin{figure}[!ht]
\begin{center}
   \includegraphics[width=0.99\linewidth]{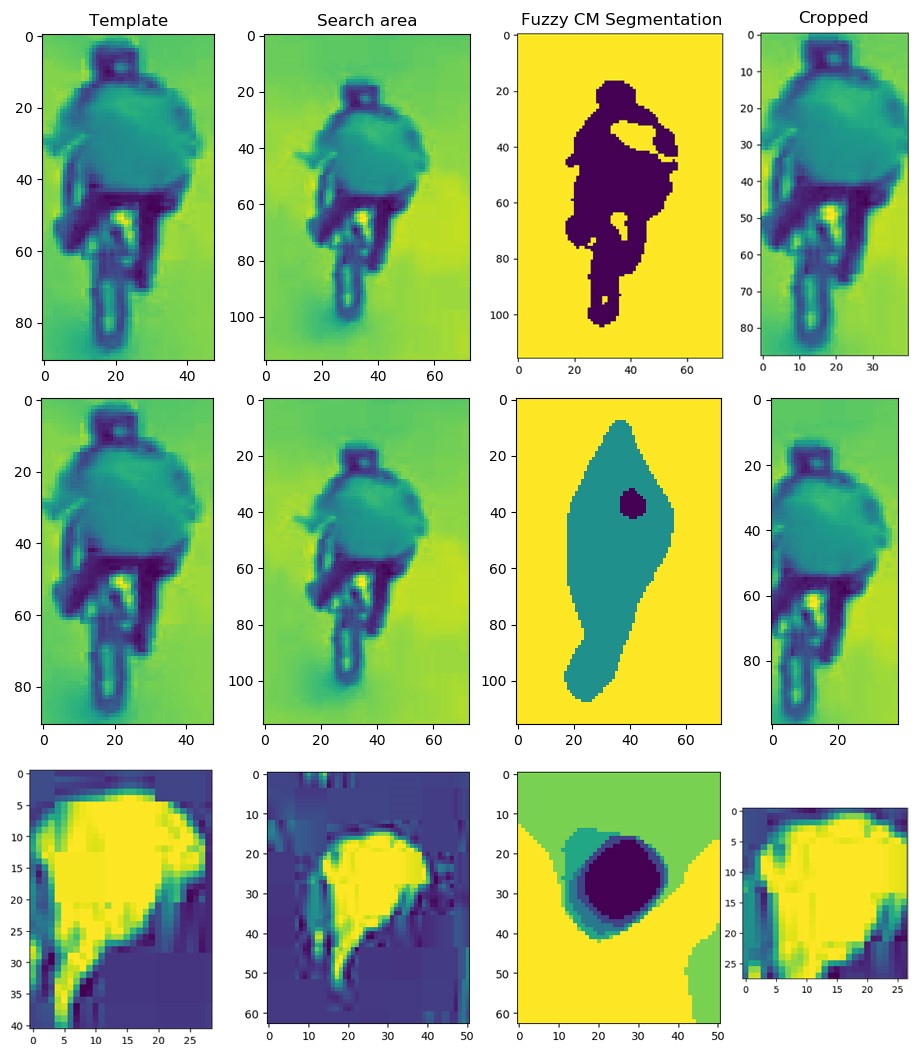}
\end{center}
   \caption{Fuzzy C Means based segmentation examples with different cluster numbers. \nth{1} and \nth{2} cluster 2 and 3 segments for the same object, and \nth{3} clusters 5 segments.}
   \label{FCM}
\label{results}
\end{figure}

We use a pre-trained VGG16 \cite{simonyan2014very} network, trained on the ImageNet dataset, to extract the convolutional features of the clustered segments. This process offers to transfer the learning of that large dataset to produce discriminative features vectors. 
Then, we calculate the cosine distance between the feature vectors of the original reference template and each segments as in Eq. \ref{cosine}. 
\begin{equation}\label{cosine}
\cos simi. ({\bf T},{\bf S})= {{\bf T} {\bf S} \over \|{\bf T}\| \|{\bf S}\|} = \frac{ \sum_{i=1}^{n}{{\bf T}_i{\bf S}_i} }{ \sqrt{\sum_{i=1}^{n}{({\bf T}_i)^2}} \sqrt{\sum_{i=1}^{n}{({\bf S}_i)^2}} }
\end{equation}
where ${\bf T}$ and ${\bf S}$ represent the template and segment feature vectors, respectively.
Fig. \ref{FCM} shows three different examples of the FCM based segmentation process. The first row shows a two-clusters FCM segmentation. This example shows better bounding box estimation that the one in the second row with three clusters. The last row comes with five clusters to extract the sheep from the surrounding grass area. We incorporate the performances for the relative angular scaling with the FCM segmentation for the best tracking results. 

\subsection{Relative Angular Scaling}\label{ars}
DroTrack updates the object scale based on its recent motion features.
Here, scale refers to the size of the current tracking template area in relation to the previous template.

The new scale is dependent on its distance from the drone camera. 
Since the utilised datasets do not have camera parameters, 
DroTrack computes the new scale in two-dimensional projection. The motion model; i.e., speed and direction, are calculated between the coordinates of the centre locations in the previous and current frames. When the object moves up in the image; i.e., has a negative change of the coordinate y, the scale is relatively reduced. Moving closer to the drone camera; i.e., having a high y value, the scale is enlarged. In order to make the scaling algorithm more accurate, we relate the scale ratio to the motion angle. The more vertical the object's direction is, the higher the scale ratio it has. For example, if an object is moving vertically at $\frac{\pi}{2}$ or $\frac{-\pi}{2}$ from the drone camera, its scale is at the highest possible ratio. However, if the moving angle becomes low, the scale ratio will be small.

\begin{figure}[t]
\begin{center}
   \includegraphics[width=0.75\linewidth, angle =0]{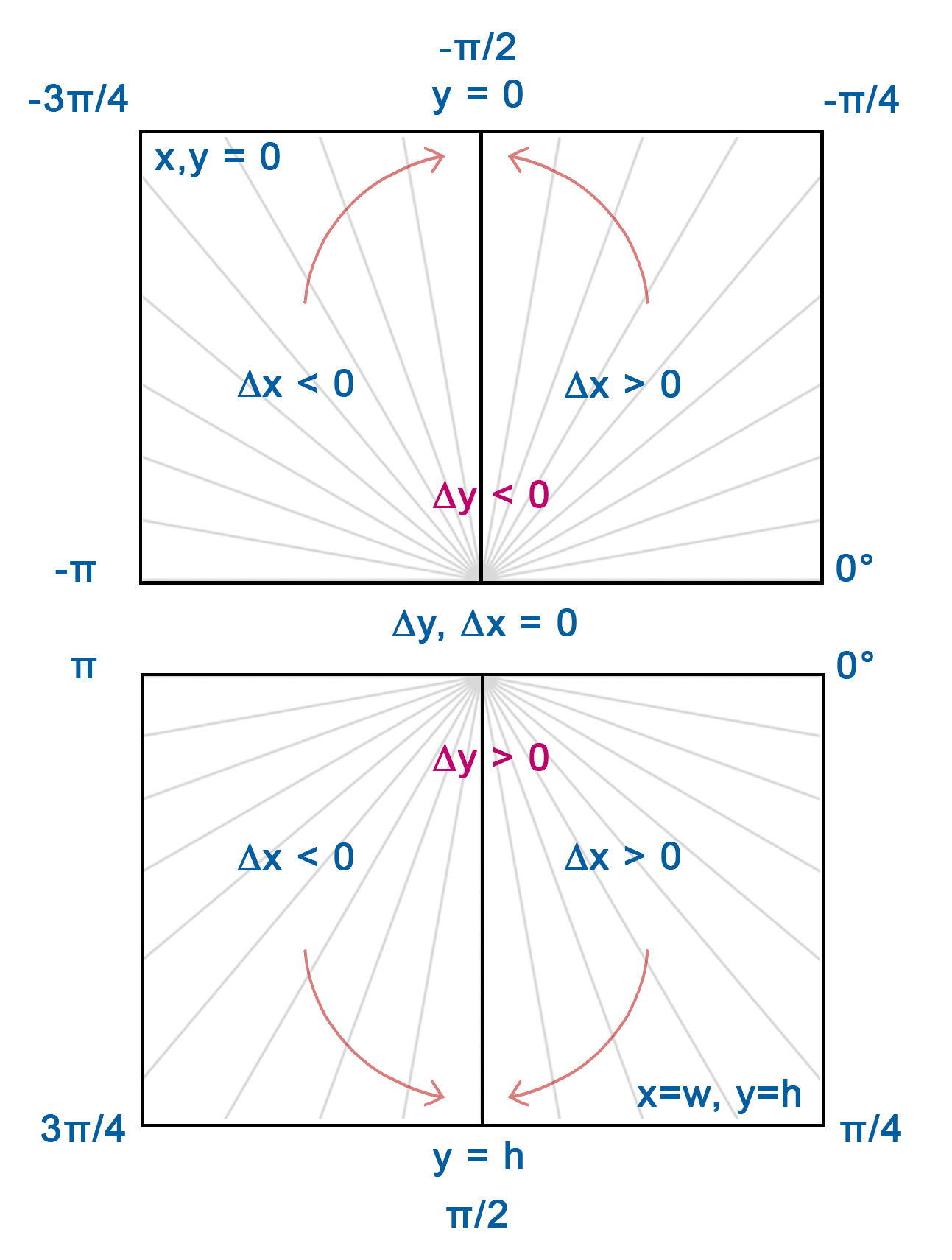}
\end{center}
   \caption{The angular relative scaling zones and their constraints. The $\Delta x$ and $\Delta y$ refer to the differences in the coordinates between the two frames.}
\label{fig:angular}
\end{figure}

Fig. \ref{fig:angular} shows the concept of the proposed relative angular scaling algorithm. We feed the algorithm with the current and previous templates as well as the motion angle ($\theta$). The angle between the coordinates x,y in the two templates is calculated using the two-argument \emph{atan2}. The \emph{atan2} computes the angle between the positive x-axis of a plane and the coordinates x, y on it according to Eq. \ref{eq:atan}.
\begin{equation}\label{eq:atan}
    \theta ((x_1,y_1), (x_2,y_2)) = atan2 (y_2 - y_1, x_2 - x_1) \in (-\pi,\pi)
\end{equation}
The frame F is divided into four zones that are sliced from the coordinates of the current location centre. In the case that $\Delta y$, i.e., the difference in the coordinate \emph{y} between the two frames is negative, two zones are defined as
$\frac{-\pi}{2} < angle < 0$ and $-\pi < angle < \frac{-\pi}{2}$. 
For the positive case, $\Delta y$, $0 < angle < \frac{\pi}{2}$ and $\frac{\pi}{2} < angle < \pi$. The red arrows in Fig. \ref{fig:angular} point up and down to the negative and positive scaling in each zone, respectively. 
The template scale is computed as its height over the height of the current frame F. The relative ratio is calculated as the ratio between the current frame \emph{y} coordinate and the previous one.
Based on the fact that the scale is relatively dependent on the motion angle, the algorithm computes the new relative scale under one of seven conditions, as in Eq. \ref{eq:rt_conditions}. In the initial case, the algorithm returns the previous RT (PRT) scale when there has been no change in the previous motion model. 
Two cases are directly scaled-down and -up for angle $\frac{-\pi}{2}$ and $\frac{\pi}{2}$, respectively. 
The other four cases include two cases when the $\Delta x > 0$ are normalised with their angle over $\frac{\pi}{2}$, and two when $\Delta x < 0$. The latter two cases have inverse directions. Therefore, they are first subtracted from 180 and normalised over $\frac{\pi}{2}$. 

\begin{equation}\label{eq:rt_conditions}
    RT = \left\{\begin{matrix} 
    PRT & \rightarrow & \Delta x \wedge \Delta y = 0; \\ 
    PRT * Scale & \rightarrow & \theta = \frac{\pi}{2} \vee \theta = \frac{-\pi}{2} ; \\ 
    PRT * Scale * \frac{\theta}{\pi/2}& \rightarrow & \Delta x > 0 \wedge \theta > 0; \\ 
    PRT * Scale * \frac{\theta}{\pi/2}& \rightarrow & \Delta x > 0 \wedge \theta < 0; \\ 
    PRT * Scale * \frac{\pi-\theta}{\pi/2}& \rightarrow & \Delta x < 0 \wedge \theta < 0; \\
    PRT * Scale * \frac{\pi-\theta}{\pi/2}& \rightarrow & \Delta x < 0 \wedge \theta > 0; \\
    \end{matrix}\right.
\end{equation}

\newlength\myindent
\setlength\myindent{2em}
\newcommand\bindent{%
  \begingroup
  \setlength{\itemindent}{\myindent}
  \addtolength{\algorithmicindent}{\myindent}
}
\newcommand\eindent{\endgroup}

The experimental results showed that the angular method is effective for scale adaptation. Since the object motion is projected in two dimensions, the algorithm can capture accurate scale changes regardless of the object direction. 

\section{Experimental Results and Evaluations}\label{exp_results}
The proposed DroTrack algorithm is implemented in Python using standard methods for feature extraction and optical flow in the OpenCV library. Fig. \ref{DroTrackFCM} highlights the work-flow and the inter-relationships among the DroTrack components. 

\begin{figure}[ht]
\begin{center}
   \includegraphics[width=.95\linewidth]{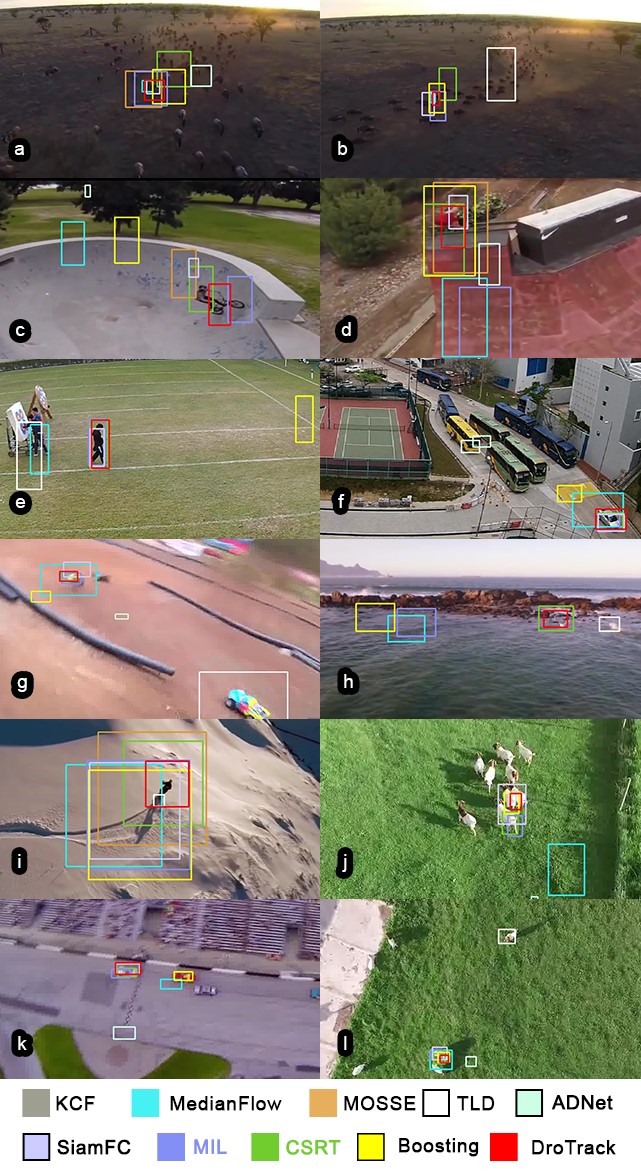}
\end{center}
   \caption{Sample frames for tracking results.}
\label{frames_results}
\end{figure}

\textbf{Datasets} We used two publicly released datasets: DTB70 \cite{li2017visual} and
UAV123 \cite{mueller2016benchmark}. 
% UAVDT-Benchmark-S \cite{du2018unmanned}, and 
% VisDrone2019-SOT \cite{zhuvisdrone2018}. 
% to evaluate the proposed tracking algorithm.
The two datasets consist of 51,462 frames. The datasets are of high diversity and captured in multiple environments. For examples, see Fig. \ref{planar}, \ref{DroTrackFCM}, and \ref{frames_results}. These datasets cover more difficulties and uncertainties aspects that are not found in the traditional tracking datasets such as VOT \cite{kristan2014visual} and VTB50 \cite{wu2013online}. The datasets include both translation and rotation camera motions. The results show that this dataset is challenging for conventional tracking algorithms. They also cover highly challenging cases in both short-term and long-term occlusion. The datasets contain different moving object types, such as humans, animals, cars, boats, birds and drones. This offers different levels of degree of freedom for the motion. Objects like cars and boats can only translate or rotate, whereas humans and animals, birds and drones have a higher degree of freedom. The datasets outdoor scenes are in various situations, including significantly varied backgrounds. These challenging motion characteristics cause object deformation, leading to more difficult object tracking.

\textbf{Evaluation metrics} To evaluate DroTrack, we computed the success overlap and centre location error. The intersection over union (IoU) is used to compute the success plots and the precision thresholds for the centre location errors. The IoU is an evaluation metric used to measure the tracking accuracy. IoU is computed for each frame using the predicted boundary box and the ground truth box. The precision score is calculated with a set of thresholds of centre location for each frame prediction. The trackers are ranked using the area under the curve (AUC) metric for the success plot and the representative precision at the threshold of ($\epsilon$ = 20 and 100) for the precision plot. All the reported results are in one-pass evaluation (OPE) settings.

\begin{figure}[!ht]
\begin{center}
   \includegraphics[width=0.75\linewidth]{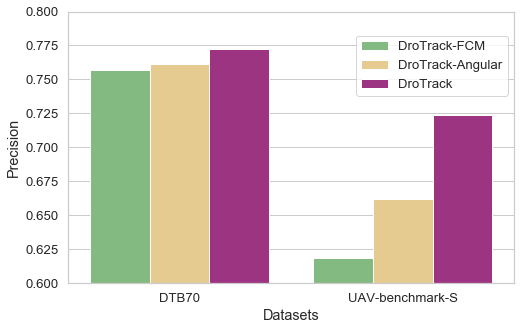}
\end{center}
   \caption{DroTrack precision of using FCM, angular geometry of for the utilised two datasets.}
\label{DroTracks}
\end{figure}

\textbf{Ablation study} We first run two versions of DroTrack (with the angular module only), on the DTB70 dataset, with and without the algorithm for the relative correction (\emph{rc}) of the optical flow. The experimental results show that the \emph{rc} algorithm improved the precision ($\epsilon=100$) score from 0.62 to 0.75 and the IoU from 0.21 to 0.25. 
For even faster tracking, we implemented DroTrack in two different modes using full-size (\emph{fs}) and half-size (\emph{hs}) frames. The results show that reducing the frame size to half decreased the tracking computational costs. However, the tracking success overlap and precision scores are slightly degraded. Table \ref{ablation_results} lists the results of the ablation study of the DroTrack versions. Using the \emph{rc} algorithm with the \emph{fs} mode produces the best centre location precision ($\epsilon = 20$) / ($\epsilon = 100$) and the best IoU success overlap score. With an average of 206 \emph{fps}, it enables these accurate results at real-time speeds. The DroTrack version without the \emph{rc} in the \emph{hs} mode results in very high speed tracking, with \textbf{1033} \emph{fps}. 
Fig. \ref{DroTracks} shows the significance of the proposed framework. It illustrates the performance precision of DroTrack using only the Fuzzy-based segmentation in comparison to the geometrical angular scaling. It also shows how DroTrack enhance the FCM based results by adding the proposed geometrical angular method. In the following experiments, we show the three different versions of DroTrack, i.e.,  DroTrack-FCM based, DroTrack-Angular, DroTrack (having both), in comparison to the state-of-the-art trackers.
\begin{table}
\caption{Comparison results between the proposed DroTrack versions using the DTB70 dataset.}\label{ablation_results}
\begin{tabular}{
|p{2.1cm}|p{.50cm}|p{.50cm}|p{1cm}|p{1cm}|p{1cm}|
}
\hline
Tracker & 
$P. 100$ & 
$P. 20$ &
IoU & 
Time & 
\emph{fps} \\ \hline

\textbf{DroTrack \emph{rc} \emph{fs}} & 
\textbf{0.75} & 
\textbf{0.41} & 
\textbf{0.25} & 
\textbf{0.0048} & 
\textbf{206} \\ \hline

\textbf{DroTrack \emph{rc} \emph{hs}} & 
\textbf{0.67} & 
\textbf{0.35} & 
0.23 & 
0.0012 & 
\textbf{840} \\ \hline

\textbf{DroTrack \emph{fs}} & 
\textbf{0.63} & 
\textbf{0.36} & 
0.21 & 
0.0036 & 
275 \\ \hline

\textbf{DroTrack \emph{hs}} & 
\textbf{0.62} & 
0.34 & 
0.20 & 
0.0010 & 
\textbf{1033} \\ \hline

\end{tabular}
Note: \emph{rc} refers to using the relative correction algorithm. \emph{fs} and \emph{hs} refer to using the full- and half-sizes of the given frames.
\end{table}

\begin{figure*}[!ht]
\begin{center}
   \includegraphics[width=0.9\linewidth]{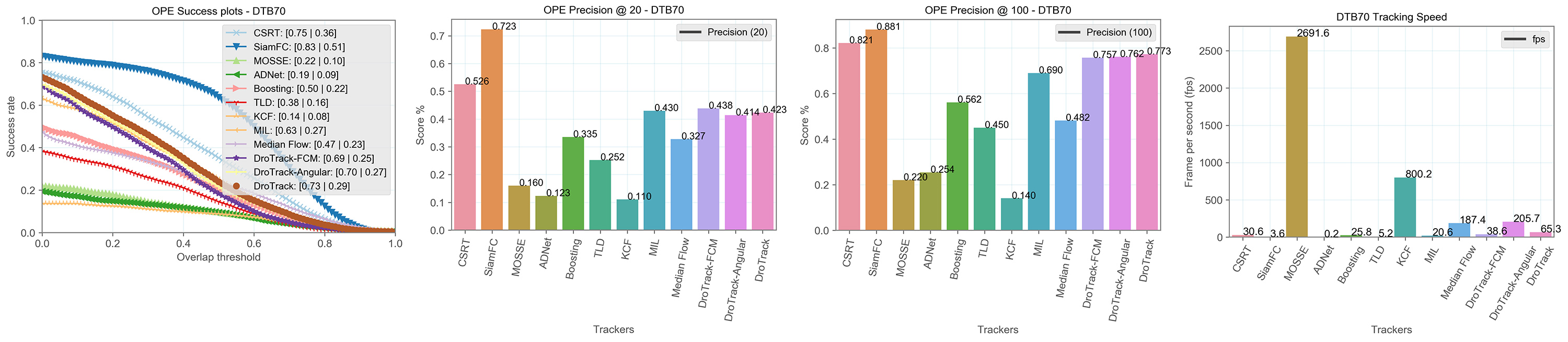}
\end{center}
   \caption{Benchmarking results of the IoU, Precision ($\epsilon = 20$) and ($\epsilon = 100$), and tracking speed \emph{fps}, for the DTB70 dataset.}
\label{results_DTB70}
\end{figure*}

\begin{figure*}[!ht]
\begin{center}
   \includegraphics[width=0.9\linewidth]{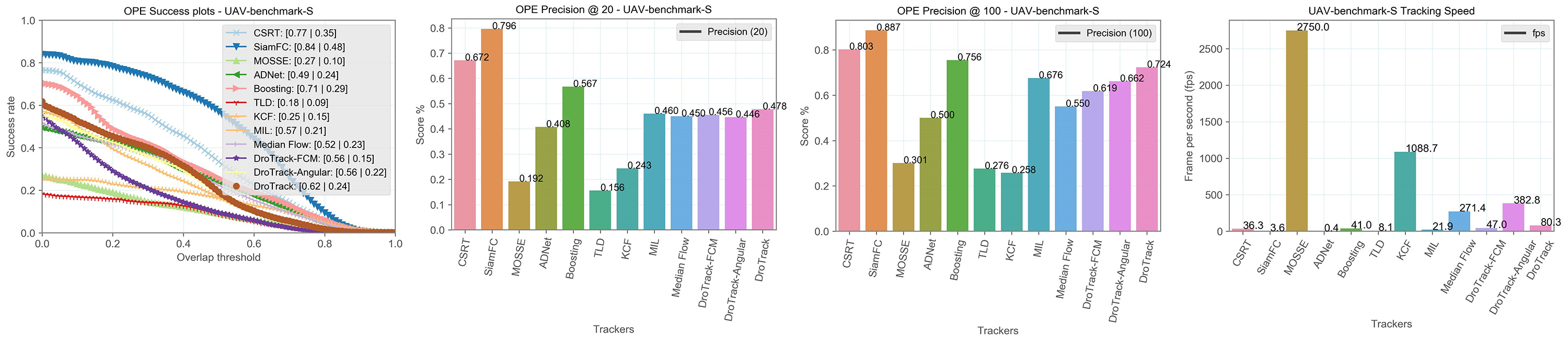}
\end{center}
   \caption{Benchmarking results of the IoU, Precision ($\epsilon = 20$) and ($\epsilon = 100$), and tracking speed \emph{fps}, for the UAV-benchmark-S dataset.}
\label{results_UAV}
\end{figure*}

\textbf{Quantitative and Qualitative Benchmark} We compare DroTrack with nine state-of-the-art and baseline trackers, including 
CSRT \cite{lukeziuaz2018discriminative}, 
ADNet \cite{yun2017action}, 
SiamFC \cite{bertinetto2016fully}, 
MIL \cite{babenko2009visual}, 
kernelised correlation filters (KCF) \cite{henriques2015high}, Median Flow \cite{kalal2010forward}, 
Boosting \cite{grabner2006real}, 
MOSSE \cite{bolme2010visual} and 
TLD \cite{kalal2012tracking}. 
The implementations of these trackers are found in OpenCV package and at the GitHub platform. 
All the benchmarking experiments were done using one CPU of Intel Core i7-3635QM and 8 GB SDRAM.
We compare three versions of DroTrack, including 1) the DroTrack with FCM, 2) DroTrack with Angular scaling, and 3) DroTrack with both FCM and Angular scaling.
Tables \ref{BM_DTB70} and \ref{BM_UAV_benchmark_S} show the mean IoU and precision, and \emph{fps} for DroTrack and the other trackers. The scores are highlighted in these tables in different colours: \textcolor{darkgreen}{\textbf{green}} for \nth{1} rank, \textcolor{blue}{\textbf{blue}} for \nth{2} rank, \textcolor{red}{\textbf{red}} for \nth{3} rank, and \textcolor{orange}{\textbf{orange}} for \nth{4} rank. Fig. \ref{results} shows the benchmarking results on the four datasets. 
The four columns compare: the IoU, Precision ($\epsilon = 20$), Precision ($\epsilon = 100$), and tracking (\emph{fps}). 
Table \ref{BM_DTB70} and Fig. \ref{results_DTB70} show the benchmarking results using the DTB70 dataset. DroTrack ranks second (out of 10) for the average distance and third for all other experiments.
Using the FCM produces good Precision results ($\epsilon = 20 \& 100$) where it ranks fourth and fifth. DroTrack with the angular method achieves better distance average and tracking speed. The combination of the two algorithms ranks second in terms of the distance average and comes better than using each method separately. Here, DroTrack outperforms all the high-speed trackers and achieves promising results in comparison to deep learning trackers. It has 0.29, 0.73 for mean and lowest IoU; 0.43 and 0.77 for the Precision; and 206 and 65 \emph{fps} for the tracking speed. 
Table \ref{BM_UAV_benchmark_S} and Fig. \ref{results_UAV} highlight the experimental benchmarking results utilising the UAV-Benchmark-S dataset. DroTrack ranks second and fourth in the average of the error distance with 82 for the combination version and 93 for the angular one, respectively. The deep learning SiamFc comes better than DroTrack with a distance error of 75. However, DroTrack still outperforms other deep learning trackers such as CSRT and ADNet. Moreover, the average tracking speed of DroTrack is high with 47, 383 (\nth{3} rank), and 80 \emph{fps} in the three versions. However, the IoU comes lower with only 0.24 and 0.62. 
% For the datasets UAV123 and VisDrones2019-SOT, Table \ref{BM_UAV123} shows that DroTrack ranks third in both centre location distance average and tracking speed while having low IoU in the UAV123 dataset. Table \ref{BM_VisDrone2019_SOT} shows that DroTrack ranks third in the distance and tracking speed. 
However, DroTrack here is still better than all the other high-speed trackers.

\begin{figure*}[!ht]
\begin{center}
   \includegraphics[width=0.9\linewidth]{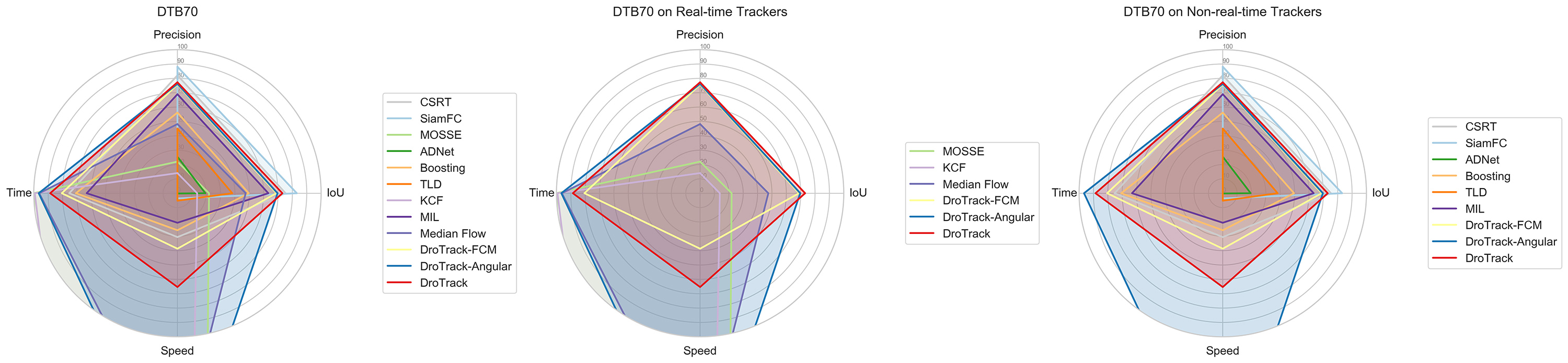}
\end{center}
   \caption{Radar charts benchmarking on the DTB70.}
\label{Radar_chart_DTB70}
\end{figure*}

\begin{figure*}[!ht]
\begin{center}
   \includegraphics[width=0.9\linewidth]{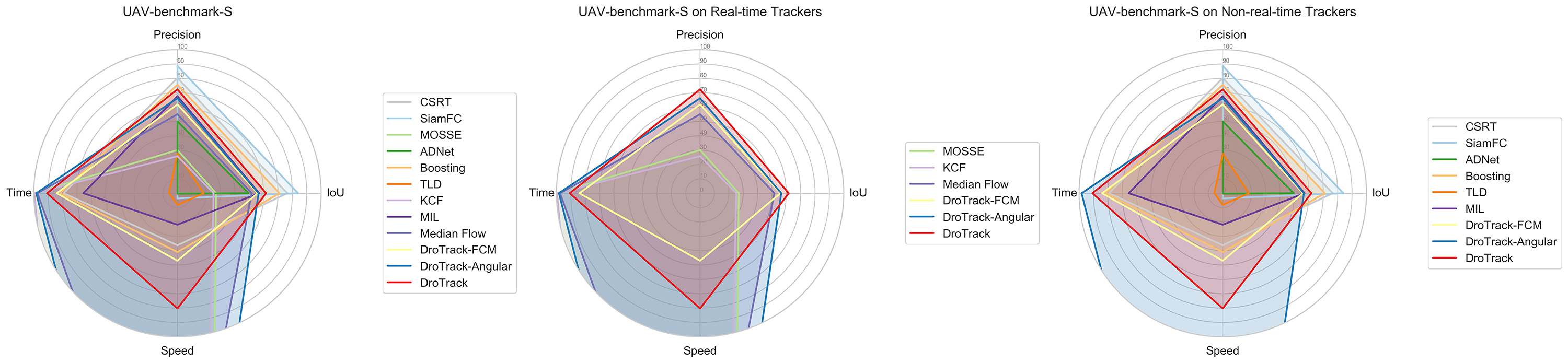}
\end{center}
   \caption{Radar charts benchmarking on the UAV-benchmark-S.}
\label{Radar_chart_UAV}
\end{figure*}

Fig. \ref{Radar_chart_DTB70} and \ref{Radar_chart_UAV} show polar radar charts comparing the three version of DroTrack with the benchmarking trackers. The charts represent the results in terms of precision, IoU, time and tracking speed. The figures show how real-time trackers produce low accuracy and high speeds. In contrast, deep learning-based trackers have high accuracy and low speeds. Here, DroTrack balances the performance between accuracy and speed. In terms of accuracy, DroTrack outperforms all the real-time trackers and compete with the deep learning ones. For the tracking speed, DroTrack outperforms all the deep learning trackers.

\begin{table}[!ht]
\caption{Benchmarking on the DTB70 dataset.}\label{BM_DTB70}
\begin{tabular}{
|p{3.2cm}|
p{.6cm}|
p{.6cm}|
p{.6cm}|
p{.6cm}|
p{.6cm}|
}
\hline
Tracker & 
$P. 100$ & 
$P. 20$ &
Dist. &
IoU  & 
\emph{fps} \\ \hline

\textbf{CSRT} \cite{lukeziuaz2018discriminative} &
\textcolor{blue}{\textbf{0.80}}&
\textcolor{blue}{\textbf{0.53}}&
\textcolor{orange}{\textbf{95}} & 
\textcolor{blue}{\textbf{0.35}} &
31 \\ \hline

\textbf{SiamFC} \cite{bertinetto2016fully} &
\textcolor{darkgreen}{\textbf{0.86}}&
\textcolor{darkgreen}{\textbf{0.72}}&
\textcolor{darkgreen}{\textbf{69}} & 
\textcolor{darkgreen}{\textbf{0.51}} &
3.6 \\ \hline

\textbf{MOSSE} \cite{bolme2010visual} & 
0.22&
0.16&
579 & 
0.10 & 
\textcolor{darkgreen}{\textbf{2692}} \\ \hline

\textbf{ADNet} \cite{yun2017action} & 
0.25&
0.12&
356 & 
0.09 & 
0.2  \\ \hline

\textbf{Boosting} \cite{grabner2006real} & 
0.55&
0.34&
184 & 
0.22 & 
25.8  \\ \hline

\textbf{TLD} \cite{kalal2012tracking} & 
0.44&
0.25&
254 & 
0.16 & 
5.2 \\ \hline

\textbf{KCF} \cite{henriques2015high} & 
0.14&
0.11&
656 & 
0.08 & 
\textcolor{blue}{\textbf{800}} \\ \hline

\textbf{MIL} \cite{babenko2009visual} &
0.68&
0.43&
131 & 
\textcolor{orange}{\textbf{0.27}} & 
21 \\ \hline

\textbf{Median Flow} \cite{kalal2010forward} & 
0.47&
0.33&
255 & 
0.23 & 
\textcolor{orange}{\textbf{187}}  \\ \hline

\textbf{DroTrack-FCM (Ours)} &
\textbf{0.74}&
\textcolor{red}{\textbf{0.44}}&
\textbf{124}       & 
\textbf{0.25} & 
\textbf{38.6} \\ \hline

\textbf{DroTrack-Angular (Ours)} &
\textcolor{orange}{\textbf{0.75}}&
\textbf{0.41}&
\textcolor{red}{\textbf{94}}       & 
\textcolor{orange}{\textbf{0.27}} & 
\textcolor{red}{\textbf{206}} \\ \hline

\textbf{DroTrack (Ours)} &
\textcolor{red}{\textbf{0.77}}&
\textcolor{orange}{\textbf{0.43}}&
\textcolor{blue}{\textbf{86}}       & 
\textcolor{red}{\textbf{0.29}} & 
\textbf{65} \\ \hline
\end{tabular}
Colours note: \textbf{
\textcolor{darkgreen}{\nth{1}}, 
\textcolor{blue}{\nth{2}}, \textcolor{red}{\nth{3}}, and \textcolor{orange}{\nth{4}} ranks.}
\end{table}

\begin{table}[!ht]
\caption{Benchmarking on the UAV-benchmark-S dataset.}\label{BM_UAV_benchmark_S}
\begin{tabular}{
|p{3.2cm}|
p{.6cm}|
p{.6cm}|
p{.6cm}|
p{.6cm}|
p{.6cm}|
}
\hline
Tracker & 
$P. 100$ & 
$P. 20$ &
Dist. &
IoU  & 
\emph{fps} \\ \hline

\textbf{CSRT} \cite{lukeziuaz2018discriminative} &  
\textcolor{blue}{\textbf{0.78}}&
\textcolor{blue}{\textbf{0.65}}&
\textcolor{red}{\textbf{89}} & 
\textcolor{blue}{\textbf{0.35}} & 
36.3 \\ \hline

\textbf{SiamFC} \cite{bertinetto2016fully} & 
\textcolor{darkgreen}{\textbf{0.86}}&
\textcolor{darkgreen}{\textbf{0.77}}&
\textcolor{darkgreen}{\textbf{75}} & 
\textcolor{darkgreen}{\textbf{0.48}} &
3.6 \\ \hline

\textbf{MOSSE} \cite{bolme2010visual} & 
0.29&
0.19&
463 & 
0.10 & 
\textcolor{darkgreen}{\textbf{2750}} \\ \hline

\textbf{ADNet} \cite{yun2017action} & 
0.49&
0.40&
236 & 
0.24 & 
0.4 \\ \hline

\textbf{Boosting} \cite{grabner2006real} & 
\textcolor{red}{\textbf{0.73}}&
\textcolor{red}{\textbf{0.55}}&
112 & 
\textcolor{red}{\textbf{0.29}} & 
41  \\ \hline

\textbf{TLD} \cite{kalal2012tracking} & 
0.27&
0.15&
298 & 
0.09 & 
8 \\ \hline

\textbf{KCF} \cite{henriques2015high} & 
0.25&
0.24&
482 & 
0.15 & 
\textcolor{blue}{\textbf{1089}} \\ \hline

\textbf{MIL} \cite{babenko2009visual} & 
0.66&
0.45&
131 & 
0.21 & 
21.9  \\ \hline

\textbf{Median Flow} \cite{kalal2010forward} &  
0.53&
0.44&
250 & 
\textcolor{orange}{\textbf{0.23}} & 
\textcolor{orange}{\textbf{271}}  \\ \hline

\textbf{DroTrack-FCM (Ours)} &

\textbf{0.60} &
\textbf{0.44} &
\textbf{188}  &
\textbf{0.15} &
\textbf{47.0} \\ \hline

\textbf{DroTrack-Angular (Ours)} &
\textbf{0.64}&
\textbf{0.43}&
\textcolor{orange}{\textbf{93}}&
\textbf{0.22}&
\textcolor{red}{\textbf{382.8}} \\ \hline

\textbf{DroTrack (Ours)} & 
\textcolor{orange}{\textbf{0.72}}&
\textcolor{orange}{\textbf{0.48}}&
\textcolor{blue}{\textbf{82}} & 
\textbf{0.24} & 
\textbf{80} \\ \hline
\end{tabular}
Colours note: \textbf{
\textcolor{darkgreen}{\nth{1}}, 
\textcolor{blue}{\nth{2}}, \textcolor{red}{\nth{3}}, and \textcolor{orange}{\nth{4}} ranks.}
\end{table}

Fig. \ref{frames_results} shows a sample of 12 frames with the predicted template rectangle for each tracker. The results show the accurate position and scale of DroTrack predictions. In many cases, such as in frames b, d, h, i, and j, the other trackers produce erroneous scales larger than the actual ones. The drone motion and scene illumination distract the trackers. However, the proposed Fuzzy segmentation and angular relative scale algorithms enable accurate DroTrack results. DroTrack's performances support this interpretation, by having low centre location errors (e.g. Precision scores) from the ground truth and consistent IoU and precision results. This low error and consistency prove the superiority of DroTrack. 
In the scene j, however, DroTrack's scale prediction is not accurate. This inaccurate prediction is due to the high similarity between the bounding box and the surrounding area. Specifically, is this scene (j), the tracked sheep is surrounded by multiple sheep. Therefore, the FCM based segmentation was not successful. However, the motion correction algorithms lead DroTrack to better locate the centre tracking than the other trackers in J. In most scenes, DroTrack seems to estimate smaller scales; however, in scene k, DroTrack has a relatively large-scale. The vertical drone motion on the moving object seems to vary the object depth significantly. In this case, if the drone moves high up from the object, the object scale will be decreased. However, the stationary nature of the object projection in two dimensions will prevent DroTrack from generating a better scale estimation. 
Therefore, it worth as future work to investigate DroTrack's sensitivity to the object's motion and try to overcome the challenge of lacking the scene depth.

\textbf{Computational performance} To compare computation time, the tracking speed columns in Fig. \ref{results_DTB70} and \ref{results_UAV} and Tables \ref{BM_DTB70} and \ref{BM_UAV_benchmark_S} summarise the averages of the execution time periods and the FPS (\emph{fps}) for each tracker. The MOSSE tracker has the best average time followed by the KCF and  half-size version of DroTrack. 
DroTrack-Angular is always faster than the other two variations. In the combined version, sometimes, the FCM segmentation process is skipped due to the poor outcome. Therefore in some cases, the speed of the combined version comes faster than the FCM based one.
Although the deep learning-based trackers, such as SiamFC achieves better IoU than DroTrack, they cannot be implemented in real-time drone scenarios. SiamFC only processed 3.5 \emph{fps} and the ADNet takes more than three seconds to process one frame. In addition to having the promising success overlap and location centre precision, DroTrack offers high frame rates of more than 1000 \emph{fps}. This high computational speed is due to the adaptive components of DroTrack; e.g., its adaptive corner detection. DroTrack starts with a high level of quality and a low number of corners to decrease the tracking time. Tracking one optimal corner contributes to the real-time performance of DroTrack. This shows the superiority of using DroTrack for high-speed real-time tracking.

\section{Conclusion}\label{conclusion}
We have introduced a novel drone-based single object tracking algorithm, called DroTrack. We described the dependency between the object motion model and the visual projection model. A Fuzzy C Means based segmentation algorithm was utilised to solve the visual uncertainty issues. An angular relative scaling algorithm was also developed to manage object scale variations. The performance of DroTrack is promising in comparison to the state-of-the-art and baseline trackers. In future work, the DroTrack scale algorithm should be enhanced to overcome the problem of the missing object depth. Although DroTrack outperforms the high-speed trackers and achieves promising results in comparison to deep learning trackers, new methods for reference template update can be considered as future work for further improvement.

\section*{Acknowledgment}
Ali Hamdi is supported by RMIT Research Stipend Scholarship. This research is also supported partially by the Australian Government through the Australian Research Council's Linkage Projects funding scheme (project LP150100246).

\bibliographystyle{unsrt}
\bibliography{DroTrack.bib}
\end{document}